# Real-time and Robust Feature Detection of Continuous Marker Pattern for Dense 3-D Deformation Measurement

*Mingxuan Li, Yen Hang Zhou, Liemin Li, Yao Jiang\**

*Department of Mechanical Engineering, Tsinghua University, 100084 Beijing, China*

**ABSTRACT**

Visuotactile sensing technology has received much attention in recent years. This article proposes a feature detection method applicable to visuotactile sensors based on continuous marker patterns (CMP) to measure 3-d deformation. First, we construct the feature model of checkerboard-like corners under contact deformation, and design a novel double-layer circular sampler. Then, we propose the judging criteria and response function of corner features by analyzing sampling signals' amplitude-frequency characteristics and circular cross-correlation behavior. The proposed feature detection algorithm fully considers the boundary characteristics retained by the corners with geometric distortion, thus enabling reliable detection at a low calculation cost. The experimental results show that the proposed method has significant advantages in terms of real-time and robustness. Finally, we have achieved the high-density (10.7 markers per $mm^2$) 3-d contact deformation visualization based on this detection method. This technique is able to clearly record the process of contact deformation, thus enabling inverse sensing of dynamic contact processes.

## 1. Introduction

Robots have increasingly expanded their applications beyond factories and become part of our daily lives [1], [2]. This trend demands higher reliability and adaptability from robots to match human dexterity and autonomous interaction abilities [3]. Currently, a promising solution is to introduce tactile perception into robot sensing systems [4], [5]. Tactile perception compensates for local contact information that cannot be obtained by robot vision approaches and can improve the interaction ability of robots in complex environments. Among the various kinds of tactile sensors that have been studied, visuotactile sensors [6], also known as vision-based tactile sensors [7] or tactile image sensors [8], have gained significant attention in recent years. Specifically, visuotactile sensors measure the contact deformation of a soft elastomer (as the contact component) by visual methods, and reconstruct tactile characteristics from the deformation information based on mapping models. Such sensors are capable of obtaining various tactile properties such as contact geometry [9], texture [10], distributed force [11], and slippage [12], and have been successfully applied in robot grasping and operation tasks [13], [14].

Compared to other types of tactile perception technologies, the core advantage of visuotactile perception is its multi-mode sensing ability. By using the deformation of soft elastomers as the input, any type of mechanical contact characteristics can be theoretically reconstructed by designing appropriate mechanical models or training mapping relationships. For example, Lepora *et al.* used CNN to extract texture features from tactile images containing deformation information [10], GelSlim 3.0 calculated the 3-d force field from the deformation field using the FEM model of the soft elastomer [11], and Sui *et al.* proposed a method for detecting incipient slip based on deformation gradient [12]. Therefore, deformation can be selected as the original tactile information of vision-based tactile sensors.

The denser the obtained deformation information, the higher the resolution of the tactile characteristics that can be reconstructed. It indicates that improving the density and completeness of deformation measurement is significant for enhancing the performance of visuotactile sensing.

The marker displacement method (MDM) [15] is a common method for visuotactile deformation measurement. A series of dot or spherical markers (i.e., marker pattern) are prepared on the inner side of the soft elastomer. Under the illumination of internal light sources, when the elastomer deforms, the movement of the markers can be captured by the built-in cameras of the sensor to form tactile images containing deformation information. This process can be seen as a discrete sampling of contact deformation by marker patterns. Besides, Li *et al.* proposed a new pattern design: continuous marker patterns (CMP) [16], [17]. CMP adopts the feature distribution with two-dimension continuity to increase the density of the original tactile information representation. Corner features in CMP have higher recognition accuracy than discrete speckles, which is similar to the performance in the planar calibration pattern that has been proved in [18]. Besides, since the rigid connection relationship (i.e., state topology) is built between the marker features in the pattern space, CMP can also improve the process stability during the marker tracking.

However, although the previous work of CMP [16] has solved the basic problem of pattern design and ensured CMP's advantages in terms of precision and reliability, the difficulty of feature detection under high density and complexity will still disturb the improvement of representation density [17]. On the one hand, as the density of markers increases, the detection algorithm's computational complexity for each image frame increases. On the other hand, the geometric distortion of dense and complex features under contact deformation is difficult to be recognized. Thus, there is still room for improvement in the extraction algorithm, especially in real-time and robustness, to achieve dense 3-d deformation sensing.

\* *Corresponding author.*
E-mail address: jiangyao@mail.tsinghua.edu.cn (Y. Jiang)

We notice that the marker patterns used in visual calibration and localization tasks and those in visuotactile sensing tasks have a certain similarity. For the feature detection of marker patterns represented by checkerboard-like patterns, there have been mature and classic works in the field of camera calibration [19]-[21]. The easy-to-detect characteristic of the checkerboard-like design makes this type of marker pattern widely used in high-precision localization. However, the existing corner detection methods are mainly applied to plane marker patterns. Wang *et al*. proposed a corner detection method for checkerboard marks on curved surfaces, but this pattern is also used for smooth plates that are not deformable [22]. In contrast, the marker pattern is always under dynamic deformation in visuotactile perception. Thus, the damage to the marker features caused by the deformation is significantly greater than the impact of camera noise, blur, and barrel distortion in visual calibration. In addition, compared with the enthusiasm for accuracy in visual calibration tasks, visuotactile sensing focuses more on real-time and reliable recognition. Due to the difference between the detection scenarios and performance requirements of the two tasks, it is necessary to customize an adaptive feature extraction method for visuotactile sensing.

This article proposes a simple and effective method for feature detection of checkerboard-like CMP, which can be used in retro-graphic sensing of dense contact deformation. This method does not rely on training data and can be directly applied to different designs of CMP-based visuotactile sensors. The main works and contributions of this article are as follows.

- We construct the feature model of the checkerboard-like CMP by comparing the visuotactile sensing tasks and the visual calibration and localization tasks.
- We design the new sampling strategy and corner detector to achieve candidate filtering, corner detection, and refinement based on the quantization processing of sampled signals.
- We use the Tac3D 3.0 experimental platform to verify the accuracy and robustness of the proposed detection method through comparative experiments. The experimental results show that compared with existing methods, the proposed method in this article has significant advantages in robustness and achieves real-time performance that meets the relevant task requirements.

The remainder of this article is organized as follows. Section 2 proposes the requirements of CMP-based visuotactile sensing through comparative analysis and offers a novel corner model and relevant sampling strategies to adapt to the new scenario. Section 3 introduces the details of the proposed detection method, mainly including sampling mode, numerical analysis of feature quantization, and the design of the rapid detector. Section 4 describes the comparative experiments evaluating the proposed method and existing approaches, especially in robustness and real-time performance. Also, the dense visualization of 3D contact deformation based on the introduced feature detection has been demonstrated. Finally, Section 5 summarizes this article.

## 2. Detection tasks and feature model

### 2.1. Visuotactile sensing tasks

Visuotactile sensing is a retro-graphic sensing technology that measures contact characteristics online. A soft elastomer is used as the contact medium, and its deformation during contact is regarded as the original tactile information to be detected. For visuotactile sensing with marker pattern, when the contact occurs, the elastomer's deformation can be characterized by the markers under the illumination of LED and collected by the camera in the form of optical signals, as shown in Fig. 1(a). Each available feature in the marker pattern is equivalent to a sampling point for deformation information. Therefore, in addition to corners, the edge lines of the checkerboard also need to be used as features to enhance the density of tactile information representation [16]. Besides, due to the demand for dense perception, it is needed to increase the density of feature information in the marker pattern of visuotactile sensing.

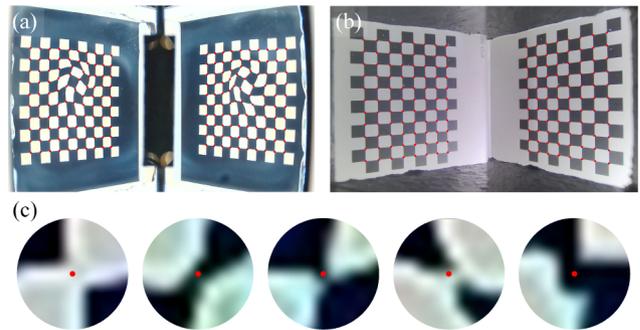

**Fig. 1.** Scenario comparison. (a) Visuotactile sensing. (b) Visual calibration and localization. (c) Examples of feature distortion that are affected by contact deformation.

Similar to visuotactile sensing, visual calibration and localization is a typical scenario that uses distinct and stable pattern features which should be robustly and accurately recognized and located by cameras. Generally, the checkerboard pattern is printed on a rigid board or projected on a screen and is guaranteed to be in the camera's field of view. By using the camera to measure the corner coordinates and organize the corner points, the spatial position and posture of the checkerboard can be obtained, as shown in Fig. 1(b). They can be used to calibrate the internal and external parameters of the camera [23] or to locate the device carrying this camera [24]. Since the purpose of visual calibration and localization tasks is to obtain precise coordinates of interest points, and using dense patterns is not necessary, only corners with prominent features are used as markers usually. Even if the edge lines of the checkerboard are detected, it is only used to refine the coordinates of the corners [22] or to assist the structure recovery [25].

Unlike visual calibration and localization, the compacted marker pattern in visuotactile sensing is always in dynamic deformation. Fig. 1(c) shows some cases of deformation effects on CMP. Since the contact tends to change the geometric shape of the pattern units, the deformed corner features no longer meet the symmetry and projection transformation relationship. Larger deformation may even cause the patterns printed on the contact surface to overlap, driving color blocks around corners to be distorted and disconnected [17]. These distortions caused by contact deformation significantly increase the difficulty of feature detection and reduce the upper limit of accuracy carried by corner features in original images. As the density of the feature increases, the size of a single marker unit decreases, and this distortion tends to be more severe. Therefore, compared with visual calibration and localization, the tasks of visuotactile sensing are more concerned with the robustness of measurement and allow the sacrifice of precision to ensure reliability in the case of large deformation.

Based on the above analysis, it can be noted that there are significant differences [see Table 1] between the two detection tasks, mainly in two aspects:

1) **Scenes and objects.**

In visual calibration and localization tasks, the checkerboard-like pattern to be detected can usually form a regular and stable image on the camera plane, which enables the frequently-used detection technologies based on the preset template [20], central symmetry [21], self-correlation [25], vertical symmetry [26], or vanishing point [27]. Even if the calibration plate is curved, the pattern's quadratic form under projective transformation can remain nondegenerate, and the corresponding approaches (such as template matching) are still applicable [22]. In contrast, the situation faced by visuotactile sensing tasks is more complex. Due to the unknown deformation mode of the marker pattern during contact, the corresponding detection algorithm needs to process the pattern image with dynamic structural deformation. These factors increase the difficulty of detection and make it hard for these prior-based methods to be effectively applied.

2) **Performance requirements.**

The nature of the task determines that the primary indicator of visual calibration and localization is accuracy. However, for visuotactile sensing, the sub-pixel quality of image features is usually not guaranteed [see Fig. 1(c)], which could reduce our expectations for the attainable accuracy of detection algorithms. In contrast, to achieve a high density of 3-d deformation representation, the real-time performance and reliability of visuotactile detection require particular attention. It means that in order to achieve dense 3D deformation measurement, the feature detection of CMP should focus on quickly and robustly obtaining a large number of corner markers with an acceptable accuracy that makes do.

The above differences indicate that the existing visual calibration and localization methods are not directly applicable to visuotactile sensing tasks. A feasible solution is to design suitable network architectures for specific vision-based sensors and achieve effective detection based on data-driven technologies [28]. However, such approaches rely on effective data samples to train the network, which could result in increased costs and reduced versatility. For simplicity and universality, this article proposes a detection method based on a feature model and sampling strategy specially designed for the visuotactile sensing task.

### 2.2. Corner Model and Sampling Strategy

#### 2.2.1 Corner model in visual calibration and localization

The distinctive characteristics of checkerboard corners in visual calibration and localization tasks make it possible to describe them using strong corner models. Different from the most "natural" corners [29], checkerboard corners meet good symmetry. Existing research used corner models to describe the pixel distribution that conforms to corners, enabling corner detectors to localize them on specific pixels. Duda et al. described corners as black and white sectors with characteristics including centerline sand crossed edges [21]. Krüger et al. [30], Yang et al. [26], and Spitschan et al. [31] defined their corner models using the angle of two intersecting edges and considered the impact of lighting and blurring on the 2-d intensity signal of a single corner. For an ideal corner model, we can summarize the following components, as shown in Fig. 2(a):

- **Symmetrical sectors (black and white)**. The pixels around a corner can be binarized and mapped to four sector regions $Z_i$ and satisfy central symmetry.
- **Two angle parameters**. Each corner is located at the intersection of the sector edges. Since the total angles of adjacent black and white sectors are always 180, only two angles $\alpha_1, \alpha_2$ are needed to describe the distribution of the four edges.

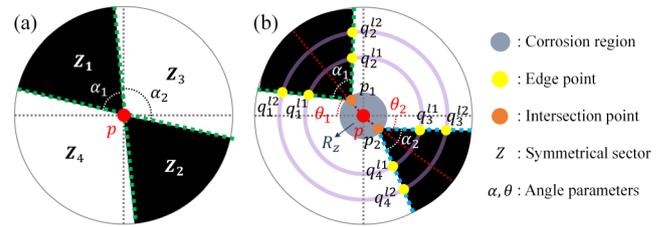

**Fig. 2.** Corner models. (a) Visual calibration and localization, (b) Visuotactile sensing based on CMP.

#### 2.2.2 Corner model in visuotactile sensing based on CMP

Section 2.1 illustrated that the distortion and error of corner images in visuotactile sensing are more significant. Compared to lighting and camera noise, the impact of contact deformation on marker pattern causes a greater deviation between real and ideal corners. In other words, using a corner model in visuotactile sensing tasks that can cover all impact factors to achieve accurate positioning is often unrealistic. Therefore, the main issue that the corner model needs to solve is to analyse imaging laws that contain light and noise rather than to define geometric features of CMP that can be maintained even under dynamic deformation.

Based on the analysis of the real corners shown in Fig. 1(c), we propose the corner model specific to visuotactile sensing based on CMP, as shown in Fig. 2(b). The geometric features in this corner model do not meet symmetry due to the compromise with contact deformation, but can still be used to describe the ideal imaging of corners in a certain deformation state. We emphasize three important components:

- **Corrosion region**. The neighborhood near each corner tends to have significant noise, distortion, and even information loss due to the impact of deformation. We define the region within a certain radius near the corner as a corrosion region $R_Z$, which means that the pixels in this area could not provide corner characteristics and localization information. The presence of corrosion regions could increase the difficulty of corner refinement.
- **Four angle parameters**. Marker patterns are usually coated or filled on a soft surface. During the deformation of soft elastomers, corners are hard to maintain connectivity and symmetry. Therefore, we need to use four angles $\alpha_1$, $\alpha_2$, $\theta_1$, $\theta_2$ to describe each boundary between black and white as the region feature. The loss of connectivity and symmetry means that corners in CMP could not be determined by the intersection of edges [32] and the saddle point of the gray-scale intensity profile [33].
- **Edge points and intersection points**. There are two special types of points on the boundary of the black and white sectors. One is the intersection $p_i$ of two adjacent boundary lines, and the other is the edge points $q_i$ that sample the edges at the same distance from the corner. Since the connecting lines between these points denote edges, they can be used to reflect the boundary feature of corners.

**Table 1**
Comparison of visual calibration & localization and visuotactile sensing tasks.

| Characteristics or requirements | | Visual calibration and localization | Visuotactile sensing based on CMP |
|---|---|---|---|
| Pattern Type | | Rigid checkerboard-like pattern | Elastic checkerboard-like pattern |
| Detection object | | Corners | Dense corners and grid-lines |
| Detection purpose | | Planar location | Contact deformation measurement |
| Detection scenario | | Unstructured environment | Structured environment |
| Requirements (relatively) | High | Accuracy | Robustness and real-time capabilities |
| | Middle | Robustness | Fineness |
| | Low | Real-time capabilities and Fineness | Accuracy |

*2.2.3 Multi-layer circular sampler*

The characteristics of the above corner model could add significant obstacles to the algorithm's implementation of visuotactile sensing based on CMP. Due to the presence of corrosion regions, corner filtering and correction cannot adapt the approaches based on pixel values (such as pixel intensity or polynomial fitting). The four angle parameters rather than two make the corner image difficult to be quantitatively described and solved. Therefore, ideas based on autocorrelation (such as symmetry) or template matching cannot be followed. The sector edges cannot maintain linear configuration during the deformation, thus making the detection based on Hough transform ineffective. Such issues could make most of the recognition and refinement techniques used in camera calibration and localization regrettably unusable.

Even if the corner does not satisfy the symmetric and junction model because of deformation, the boundary feature defined by the edge lines is able to remain identifiable and stable. One valuable idea for such requirements is using a sampling circle similar to the FAST detector [34]. For detectors designed for checkerboard-like patterns, existing research has focused on the grayscale-changing characteristic of the circular sampling boundaries. Sun *et al.* used rectangular windows to transform the 2-d pixel distribution near each corner into a 1-d binary vector [35]. Bennett *et al.* proposed the single-layer ChESS detector, which can detect corner features using the overall response values [36]. Bok *et al.* used sign-changing indicators to determine corners in the dense single-layer sampling circle [37]. Zhang *et al.* described pixel intensity distribution in the ChESS detector as a square wave with two periods [38].

The circular sampling mode eliminates redundant information within the black-and-white sectors and focuses directly on the boundary feature in corner models, which is particularly suitable for visuotactile sensing. On this basis, we further emphasize the following three sampling strategies:
- **Multi-layer sampling**. Under deformation, the edge images appear as curves, thus a single-layer sampler may not fully reflect the boundary feature.
- **Equidistant sampling.** It ensures rotational invariance for detection under camera projective transformation.
- **Continuous sampling.** It ensures that the sampling vector covers every orientation around each corner.

Based on the above sampling strategy, this article designs the corresponding samplers for diagonal filtering and refinement steps, and implements a vision-based tactile sensing approach based on CMP through topology connection, pattern recovery, and matching. The details are introduced in Section 3.

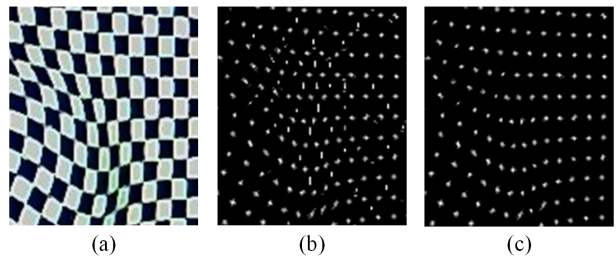

**Fig. 3.** Comparison of binarization schemes. (a) Original image. (b) Result of layer-scope threshold method. (c) Result of adaptive threshold method.

## 3. Methodology

*3.1. Approximate condition*

Corner filtering judges whether a pixel belongs to a corner based on related quantitative indicators (determined by the corner model), finds the pixel regions that may be corner points, and adds them to the candidate queue. According to the analysis in Section 2, the quality of corners in visuotactile sensing is poor, and therefore the screening criteria need to be relaxed to improve the robustness of detection. To compromise the characteristics of the ideal corner model to the actual situation, we introduce two approximate conditions as the basis for sampler design:
1) **Binary form.**

We recommend using binarization to process images and classifying all pixels as white or black (mapping the grayscale value to 255 or 0). The intention of this approximation lies in ignoring the pixel value of the corner image but paying attention to the boundary features at the boundary of the black and white sectors. This process could improve the sensitivity to low-quality corner points at the expense of localization accuracy.

In existing studies, layer-scope threshold methods were used to reduce noise in the circular boundary vector. For example, Sun *et al.* used the mean value of pixels in the vector as the threshold [35], while Bok *et al.* used the average of the maximum and minimum pixels in the vector as the threshold [37]. However, layer-scope threshold methods have poor noise suppression under uneven color conditions. The 1-d image binarization ignores pixels in the domain of corner candidates that do not belong to the loop, thereby amplifying the negative effects of noise on the jump characteristics. In contrast, the black-and-white pattern format highlights the detail retention effect of CMP in 2-d image binarization. As shown in Fig. 3, our brief evaluation show that using adaptive threshold binarization to process images directly can achieve cleaner results (using the proposed corner filtering method introduced in Section 3.4).
2) **Integer form.**

We recommend sampling and calculating only at the pixel level to reduce the amount of unnecessary computation. Bok *et al.* extracted a dense circular boundary vector around the corner candidate, and the coordinates of the samples were calculated by bilinear interpolation at the sub-pixel level [37]. This approach can improve accuracy in iterative refinement. However, the large loss of original information in distorted corner images makes it difficult to enhance measurement by calculating subpixel sampling points. A pixel-level approximation is necessary to consider the trade-off of computational overhead and revenue.

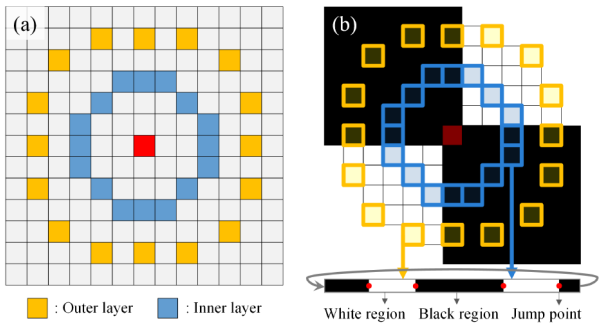

**Fig. 4.** Sampler for corner filtering. (a) Sampler design. (b) Sampling mode, two coupled boundary vectors.

### 3.2. Sampler strategy and design

The core of corner filtering is the double-layer circular sampler shown in Fig. 4(a). We adopt the sampling mode with a radius of 3 pixels for the inner layer, similar to the Fast detector [34]. For the outer layer, we use a design similar to the ChESS detector with a radius of 5 pixels [36]. In each layer of the proposed sampler, it is possible to obtain sampling pixels that are approximately consistent with the distance and angular distance from the center, equivalent to a 1-d circular sequence of 16 pixels in length. Such sampling mode can downsample the 2-d intensity distribution in the corner model into two sets of 1-d image signals [see Fig. 4(b)].

Note that the two sampling circles are used simultaneously in the corner filtering process. Bennett *et al.* believed that since the inner circle was closer to the corrosion region, it could provide almost no beneficial response compared to the outer ring. Therefore, it is appropriate only to use a single circle [36]. This conclusion is correct for visual calibration and localization tasks. However, the analysis in Section 2 has shown that the lack of connectivity and symmetry in the corner image could cause deviations in the responses obtained from different layers in visuotactile sensing tasks. Besides, contact deformation may not necessarily impact the inner layer more than the outer layer. In other words, adopting the multi-layer sampling strategy can adapt to more stringent detection conditions by introducing boundary features that are less affected by geometric distortion. Based on this analysis, our design is closer to the approach of Sun *et al.* [35]. Inspired by Bennett *et al.* [36], we try to improve image processing speed by selecting the smallest possible number of sampler layers (two layers).

The proposed double-layer sampler can also meet the other two sampling strategies (equidistant and continuous) mentioned in Section 2.2.3. For equidistant sampling, although it is not guaranteed that each sampling point has the same distance from the center, the proposed circular structure can already obtain acceptable sampling results in a discrete pixel grid. This property can be explained by Bresenham's algorithm [39]. If different sizes of circles are needed depending on the optical system used, we also recommend designing the sampler according to the pixel arrangement of the Breseham circle. For continuous sampling, the inner layer of the Fast-like sampler adopts a complete circular path, while the outer layer of a ChESS-like sampler has gaps. It has been explained in [36] that a ChESS-like sampler can already meet the optimal partition spacing, and adding additional samples will increase unnecessary computational complexity. In addition, the two sampling layers have the same length of sequence, which allows us to define the inter-layer feature response by directly comparing the similarity of two sets of vectors. Such information cannot be obtained by single-layer sampling.

The size of the sampler we choose depends on the requirement of visuotactile density characterization. If the size is too small, the sampled pixels may fall into the corrosion region, making it unable to respond to the boundary feature of the corners. Besides, when the size is too large, the sampling range may overlap with other marker units, resulting in incorrect responses. Therefore, the inner and outer layers have different tendencies during the detection: the inner layer is the benchmark and main executor of corner filtering. It has a small size (3px) and is close to the corrosion area, and requires strict evaluation to prevent misjudgments. The outer layer is larger (5px) and is more likely to overlap with other contours. Thus, a more relaxed boundary feature evaluation standard is needed to reduce false corners from noise.

In addition, like other circular templates, our sampler adopts a fixed size and does not have scale invariance. An inspiring solution is using the image pyramid scheme similar to the ORB algorithm [40]. By setting a scaling factor and the number of layers of the pyramid, the original image is reduced or enlarged to multiple images according to the scaling factor. In these images of different proportions, feature points are extracted, and the set of results with the closest number of feature points to the true value is used as the final candidate corner points. Although we did not use it in application since the proposed sampler could already meet the requirements, this strategy in other optical systems can be considered.

### 3.3. Feature quantification

Section 3.2 determines the sampling mode of the double-layer circular ring, which means that the available image information is simplified into two sets of binarized 1-d circular sequences [see Fig. 5 (a)]. On this basis, the main idea for judging corner features is to find quantitative indicators that comply with the corner model proposed in Section 2.2. Such indicators need to have the following characteristics:

- They are less affected by perspective projection transformation and contact deformation.
- They can effectively distinguish corners from other features.
- The response function corresponding to the indicators is positively correlated with the intensity of corner features.

In this article, we propose both intra-layer criterion and inter-layer criterion. The intra-layer standard is used to quantify the corner features described by the boundary features of the sampled signals within the same layer, and the response function is determined by the amplitude-frequency characteristic of the sequence [see Fig. 5 (b)]. The inter-layer criterion describes the corner features defined by the relationship between the two-layer sampling sequence, and we use the analysis method of circular cross-correlation to provide the judgment [see Fig. 5 (c)]. The proposed response functions can effectively express the characteristic attributes that CMP corners can maintain during contact deformation.

#### 3.3.1 Intra-layer criterion

Consider an ideal CMP corner model according to Section 2.2. Take a corner as the center, and sample clockwise on a circle. As shown in Fig. 6 (a), a continuous sampling signal $f(\theta)$ can be obtained. According to the characteristics of the corner model, the definition domain length of $f(\theta)$ is $2\pi$, the value is 1 or 0 (take the black value as one and the white value as zero), and $f(\theta)$ contains two rectangular window signals with unequal widths. Without loss of generality, we take the midpoint of the first rectangular window as the origin. Let the widths of window signals be $2\tau_1$ and $2\tau_2$, respectively, which meet the requirements

$$0 < \tau_1 < \frac{\pi}{2}, 0 < \tau_2 < \frac{\pi}{2}, 0 < \alpha < \pi. \quad (1)$$

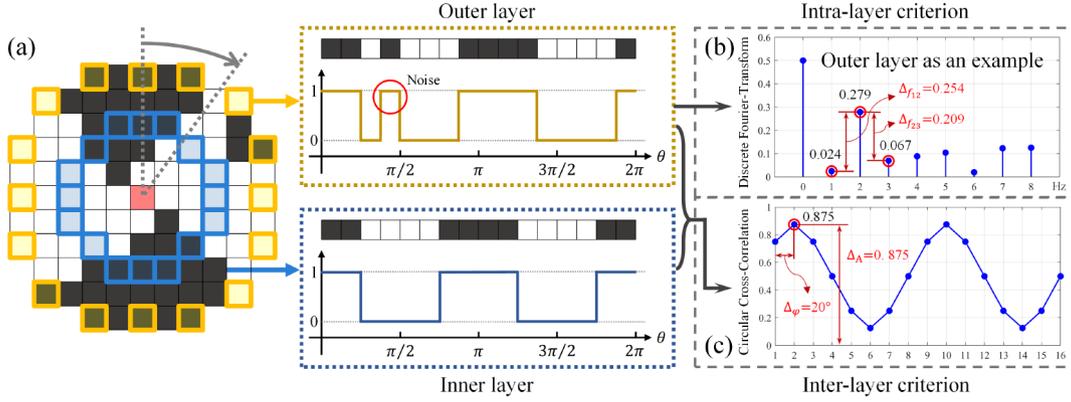

**Fig. 5.** Feature quantization. (a) Sampling sequence of the double-layer circular vectors. (b) Intra-layer criterion. (c) Inter-layer criterion.

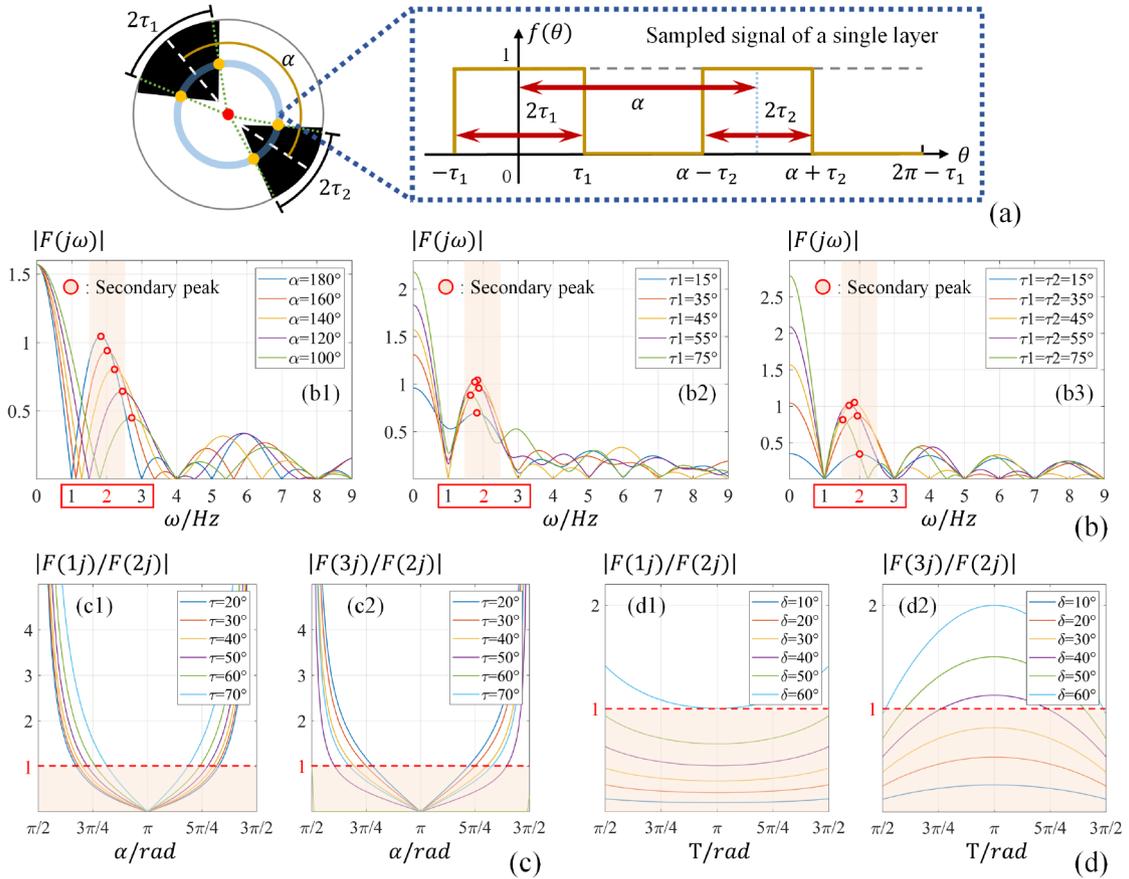

**Fig. 6.** Numerical analysis of the intra-layer sequence. (a) A function expression of the sampled signal. (b) The amplitude intensity variation of sampled signals at different frequencies. (c) Simplification: only the orientation deviation of black sectors is considered. (d) Simplification: only the length deviation of black sectors is considered.

Therefore,

$$f(\theta) = u(\theta + \tau_1) - u(\theta - \tau_1) + u(\theta + \tau_2 - \alpha) \quad (2)$$

$$-u(\theta - \tau_2 - \alpha),$$

where $u$ denotes the Step Function, and

$$u(\theta - \theta_0) = \begin{cases} 1, & \theta \geq \theta_0 \\ 0, & \theta > \theta_0 \end{cases}. \quad (3)$$

We perform the Fourier transform on $f(\theta)$, and obtain the frequency spectrum function

$$F(j\omega) = \int_{-\infty}^{+\infty} f(\theta) \cdot e^{-j\omega t} = \tau_1 \cdot Sa(\omega\tau_1) \\ + \tau_2 \cdot Sa(\omega\tau_2) \cdot e^{-j\omega\alpha}, \quad (4)$$

Where $Sa$ represents the Sampling Function. Therefore,

$$|F(j\omega)| = \omega^{-1} \cdot \\ \sqrt{sin(\omega\tau_1)^2 + sin(\omega\tau_2)^2 + 2\,sin(\omega\tau_1)\,sin(\omega\tau_2)\,cos(\omega\alpha)}. \quad (5)$$

Equ. (5) describes the relationship between the amplitude, frequency, and corner parameters of the circular sampling signal. For the corner points with the most obvious features (orthographic projection and no geometric distortion), $\tau_1 = \tau_2 = \pi/4$ and $\alpha = \pi$ should meet. For actual corners, although contact deformation could cause some deviation in the corner parameters, it should still be within a reasonable range. For example, when $\tau_1$ or $\tau_2$ is too small, the sampling center should be judged as a false corner caused by noise and removed.

As shown in Fig. 6 (b), we change the values of $\tau_1$, $\tau_2$, and $\alpha$ to analyse the amplitude intensity variation of the sampling signal at different frequencies. The amplitude diagram superimposed by the two window functions is presented as a multi-level wave peak from 0Hz to the flank. Note that as the parameters change, the abscissa of the secondary peak can be maintained within the range of 1Hz to 3Hz:

- As shown in Fig. 6(b1), fix $\tau_1 = \tau_2 = \pi/4$ and change the value of $\alpha$. This set of parameters is selected to represent the impact of the relative azimuth deviation of sectors on the amplitude intensity (using black sectors as a reference unless otherwise mentioned). When the deviation between $\alpha$ and the ideal value $\pi$ does not exceed $\pi/3$, the abscissa of the secondary peak is stable within the 1-3Hz range and close to 2Hz (i.e., the yellow region). It means that the amplitude at 2Hz is greater than that corresponding to 1Hz and 3Hz. Besides, as $\alpha$ deviates from the ideal value $\pi$, the abscissa of the secondary peak deviates from the 2Hz, and the ordinate decreases. It manifests as the difference between the amplitude at 2Hz and amplitudes at 1Hz or 3Hz gradually reducing.
- As shown in Fig. 6(b2), fix $\tau_2 = \pi/4$, $\alpha = \pi$, and change the value of $\tau_1$. This set of parameters is selected to represent the impact of the relative length deviation of sectors on the amplitude intensity. When the deviation between $\tau_1$ and $\tau_2$ does not exceed $\pi/3$, the above relationship also satisfies.
- As shown in Fig. 6(b3), fix $\tau_1 = \tau_2 = \tau$, $\alpha = \pi$, and change the value of $\tau$. This set of parameters is selected to represent the impact of the length of sectors on the amplitude intensity. The above change trend still holds true when the deviation between $\tau$ and the ideal value $\pi$ does not exceed $\pi/4$.

The above numerical analysis results indicate that the performance of secondary peaks in the sampled signal's amplitude map is enlightening for corner judgment. The closer the secondary peak is to the 2Hz position on the horizontal axis (and the larger the vertical axis value), the more the sampling center conforms to the corner features. Therefore, we can use the D-value of amplitude intensity

$$\Delta_{f_{12}} = |F(2j)| - |F(1j)|, \\ \Delta_{f_{23}} = |F(2j)| - |F(3j)|, \quad (6)$$

as the intra-layer corner response. The purpose of choosing difference instead of ratio here is to avoid divergence. As the value of corner features related to the intra-layer criterion decreases, $\Delta_{f_{12}}$ and $\Delta_{f_{23}}$ also decrease, and could be less than 0 when the feature response is low to a certain extent. It means that selecting a reasonable threshold can exclude candidate points with lower corner features, thus effectively distinguishing strong feature points from false response corners generated by noise.

We further analyse the rationality of $\Delta_{f_{12}}$ and $\Delta_{f_{23}}$, and the threshold selection by considering the sampling results under two simplified scenarios:

1) **Only consider azimuth deviation.**

Let $\tau_1 = \tau_2 = \tau$. According to Equ. (5),

$$|F(j\omega)| = \left|2\omega^{-1} sin(\omega\tau) cos\left(\frac{1}{2}\omega\alpha\right)\right|, \quad (7)$$

where $0 < \tau < \pi/2$ and $0 < \alpha < \pi$. And therefore,

$$\left|\frac{F(1j)}{F(2j)}\right| = \left|\frac{2\,sin(\tau)\,cos\left(\frac{1}{2}\alpha\right)}{sin(2\tau)\,cos(\alpha)}\right|, \quad (8)$$

$$\left|\frac{F(3j)}{F(2j)}\right| = \left|\frac{2\,sin(\tau)\,cos\left(\frac{1}{2}\alpha\right)}{3\,sin(2\tau)\,cos(\alpha)}\right|, \quad (9)$$

The function images of equations (8) and (9) are shown in Fig. 6(c). The yellow region with a vertical coordinate less than 1 represents $\Delta_{f_{12}} > 0$ or $\Delta_{f_{23}} > 0$. It can be seen that when the deviation between $\tau$ and the ideal value $\pi/4$ is within a reasonable value (approximately $\pi/9$), the values of $\left|\frac{F(1j)}{F(2j)}\right|$ and $\left|\frac{F(3j)}{F(2j)}\right|$ are relatively less affected by $\tau$. At the same time, when the deviation of $\alpha$ from the ideal value $\pi$ exceeds approximately $\pi/4$, the values of $\left|\frac{F(1j)}{F(2j)}\right|$ and $\left|\frac{F(3j)}{F(2j)}\right|$ exceed 1 and increasing rapidly. The above results indicate that the selected response quantity is sensitive to the azimuth deviation $\alpha$, but has a higher tolerance for changes in the overall length of black sectors. This selective difference helps to suppress candidate corners far from the center (within the white sector) while retaining the true corners that have become "thin" or "fat" in their sector due to stretching deformation.

2) **Only consider length deviation.**

Let $\alpha = \pi$. Without loss of generality, set $\tau_1 \geq \tau_2$. From Equ. (5),

$$|F(j\omega)| = \begin{cases} \omega^{-1} \cdot |sin(\omega\tau_1) - sin(\omega\tau_2)|, & \omega \text{ is odd} \\ \omega^{-1} \cdot |sin(\omega\tau_1) + sin(\omega\tau_2)|, & \omega \text{ is even} \end{cases}, \quad (10)$$

where $0 < \tau_2 < \tau_1 < \pi/2$. Define $T = \tau_1 + \tau_2$ and $\delta = \tau_1 - \tau_2$, and thus

$$\left|\frac{F(1j)}{F(2j)}\right| = \left|\frac{sin\left(\frac{1}{2}\delta\right)}{sin\left(\frac{1}{2}T\right)cos(\delta)}\right|, \quad (11)$$

$$\left|\frac{F(3j)}{F(2j)}\right| = \left|\frac{2\,cos\left(\frac{3}{2}T\right)sin\left(\frac{3}{2}\delta\right)}{3\,sin(T)\,cos(\delta)}\right|, \quad (12)$$

The function images of equations (8) and (9) are shown in Fig. 6(c). The yellow region with a vertical coordinate less than 1 represents $\Delta_{f_{12}} > 0$ or $\Delta_{f_{23}} > 0$. It can be seen that $T$ has little impact on the values of $\Delta_{f_{12}}$ and $\Delta_{f_{23}}$, but the values of $\left|\frac{F(1j)}{F(2j)}\right|$ and $\left|\frac{F(3j)}{F(2j)}\right|$ exceed 1 and increasing rapidly when the deviation between $\delta$ and the ideal value 0 is within a reasonable value (approximately). It means that the selected response quantity is also sensitive to the length deviation $\alpha$. This selective difference helps suppress candidate corners far from the center (within the black sector) while retaining the true corners as above. When $\Delta_{f_{12}} = 0$ and $\Delta_{f_{23}} = 0$ are selected as the thresholds, candidate points with low confidence that have a length deviation approximately $\pi/4$ can be reliably filtered.

The above discussion proves that use $\Delta_{f_{12}}$ and $\Delta_{f_{23}}$ as the intra-layer response variables of corners (such as requiring $\Delta_{f_{12}} > 0$ and $\Delta_{f_{23}} > 0$) can effectively describe the characteristics of ideal CMP corners. For images that have already been binarized, when the center of the sampler falls on other types of features, it would exhibit vastly different response. When the center is located on the edge line, the spectrum amplitude at 1Hz is always greater than the amplitude at 2Hz (i.e., $\Delta_{f_{12}} < 0$) [38]. When the center is located in a uniform area, the spectrum amplitude is all 0 at 1Hz, 2Hz, and 3Hz positions. Although uneven lighting and noise exist in actual digital images, the 3Hz amplitude tends to be greater than the 1Hz amplitude since the amplitude corresponding to higher frequencies has a greater response to noise (i.e., $\Delta_{f_{23}} < 0$).

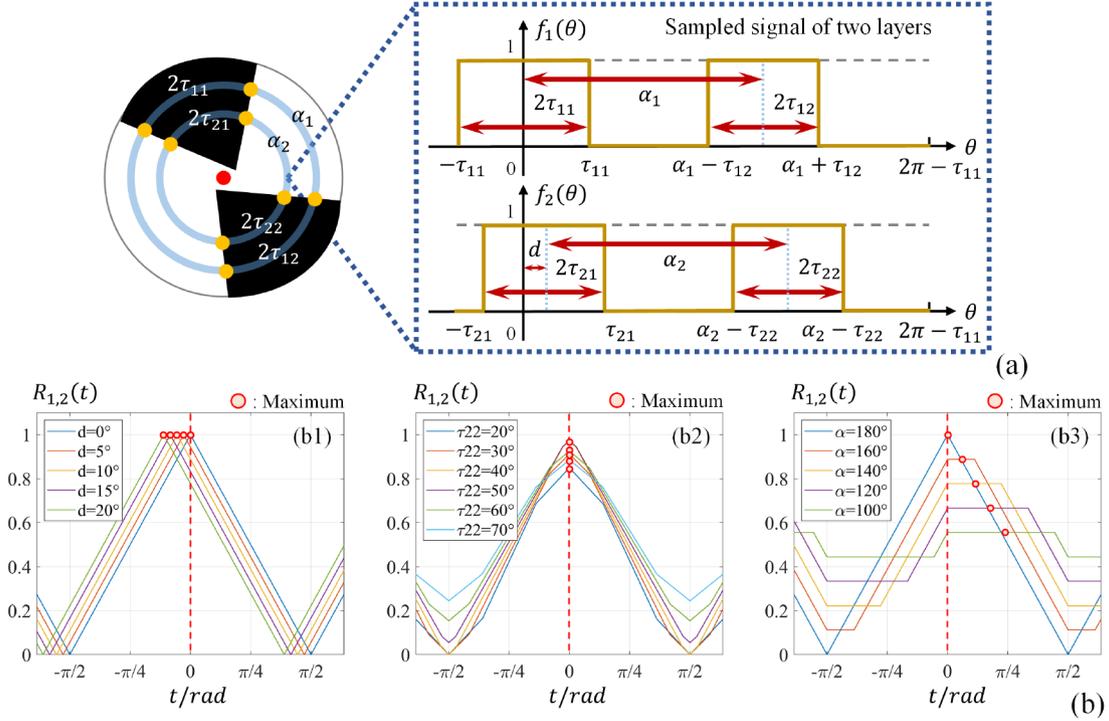

**Fig. 7.** Numerical analysis of the inter-layer sequence. (a) A function expression of the sampled signal. (b) The amplitude intensity variation of sampled signals at different frequencies, using different parameters.

For the discrete sampler in actual use, the sampling signal can be represented as a sequence $g(i)$ with a length of 16, and each element represents the grayscale value of a sampling point (after binarization). By using discrete Fourier transform (DFT), the spectral characteristics can be expressed as

$$|F(kj)| = \left|\sum_{i=0}^{15} g(i) \cdot W_{16}^{ik}\right|, \quad (13)$$
$$\text{where } W_{16} = e^{-j\frac{\pi}{8}}, \quad k \in [0, 15],$$

and $|F(kj)|$ denotes the amplitude of the sampled signal at kHz. The intra-layer response value of corners can still be defined as $\Delta_{f_{12}} = |F(2j)| - |F(1j)|$ and $\Delta_{f_{23}} = |F(2j)| - |F(3j)|$. Besides, according to Fig. 6 (c) and 6 (d), $\Delta_{f_{12}} = |F(2j)| \cdot (|F(2j)| - |F(1j)|)$ and $\Delta_{f_{23}} = |F(2j)| \cdot (|F(2j)| - |F(3j)|)$ can also be selected to increase the sensitivity of the response values. It also helps to improve the suppression of false corners from noise in smooth regions.

*3.3.2 Inter-layer criterion*

Take a corner as the center, and sample clockwise on a double-layer circumference, two sets of continuous signals $f_1(\theta)$ and $f_2(\theta)$ can be obtained as shown in Fig. 7(a). Section 2.2.2 mentions that the corner model of CMP-based visuotactile sensing does not has connectivity and symmetry. Therefore, there may be some deviation in the function images of $f_1(\theta)$ and $f_2(\theta)$. Without loss of generality, we take the midpoint of first rectangular window of $f_1(\theta)$ as the origin. Let the midpoint position of the first rectangular window of $f_2(\theta)$ be $d$. Let the widths of window signals be $2\tau_{11}$ and $2\tau_{12}$, respectively, and their center-to-center distance be $\alpha_1$. Let the relevant parameters in $f_2(\theta)$ be $2\tau_{21}$, $2\tau_{22}$, and $\alpha_2$. Therefore, the sampled signal can be represented as

$$\begin{aligned} f_1(\theta) &= u(\theta + \tau_{11}) - u(\theta - \tau_{11}) + u(\theta + \tau_{12} - \alpha_1) \\ &\quad - u(\theta - \tau_{12} - \alpha_1), \\ f_2(\theta) &= u(\theta' + \tau_{21}) - u(\theta' - \tau_{12}) + u(\theta' + \tau_{22} - \alpha_2) \\ &\quad - u(\theta' - \tau_{22} - \alpha_2), \\ &\text{where } \theta' = \theta - d. \end{aligned} \quad (14)$$

For the corners with the most obvious features (orthographic projection and no geometric distortion), $f_1(\theta)$ and $f_2(\theta)$ should be equal. The more significant the degree of geometric distortion, the greater the difference between $f_1(\theta)$ and $f_2(\theta)$. This property inspires us to use the circular cross-correlation function to describe the inter-layer relevance of double-layer sampled signals:

$$R_{1,2}(\varphi) = \int_{-\infty}^{+\infty} f_1^*(\theta) \cdot f_2^*(\theta - \varphi) \, d\theta, \quad (15)$$

where $f_1^*(\theta)$ and $f_2^*(\theta)$ represent the periodic extension of $f_1(\theta)$ and $f_2(\theta)$ (the extended intervals are the definition domains of two signals, respectively), and $R_{1,2}(\varphi)$ represents the circular cross-correlation function between sampled signals after periodic extension. Since $f_1^*(\theta)$ and $f_2^*(\theta)$ are all real functions, $R_{1,2}(\varphi) = R_{2,1}(-\varphi)$ always hold, which means relying solely on $R_{1,2}(t)$ can describe the similarity between two sets of periodic sampled signals.

In an ideal situation, the angle parameters should meet $\tau_{11} = \tau_{12} = \tau_{21} = \tau_{22} = \pi/4$, $\alpha_1 = \alpha_2 = \pi$, and $d = 0$. At this point, the maximum value of the circular cross-correlation function (i.e., the cross-correlation coefficient) is located at $\varphi = 0$, and the minimum value is located at $\varphi = \pm\pi$. As shown in Fig. 7(b), we change the values of some angle parameters and analyse the variation of the circular cross-correlation function:

- As shown in Fig. 7(b1), fix other parameters and change the value of $d$. As $d$ increases, the cross-correlation coefficient gradually deviates from the coordinate origin, and the variation of the phase coordinate is proportional to $d$. It means that the phase shift at the maximum position can be used to describe the impact of the overall shift between two sets of signals on circular cross-correlation. When the phase shift reaches a certain level (i.e., the misalignment of signals is significant), the sampler may span different marker units at this time, and the sampling center should not be considered a true corner.
- As shown in Fig. 7(b2), fix other parameters and change the value of $\tau_{22}$. When $\tau_{22}$ increases, the cross-correlation coefficient decreases accordingly. Therefore, the cross-correlation coefficient can be used to describe the impact of length deviation. When the reduction reaches a certain level (i.e., the similarity of signals is weak), the sampler may locate at a uniform area with noise, and the sampling center should not be considered a true corner.
- As shown in Fig. 7(b3), fix other parameters and change the value of $\alpha$. As the value of $\alpha$ increases, the phase shift and amplitude decrease at the maximum cross-correlation value coexist. In this case, the symmetry and similarity between the two sampled signals can be described by phase shift and cross-correlation coefficient, respectively.

The above discussion proves that the phase shift and cross-correlation coefficient of the circular cross-correlation function can be used to describe the inter-layer corner response, which can be expressed as

$$\Delta_A = R_{1,2}(\varphi_{max}),$$
$$\Delta_\varphi = \left|\frac{\varphi_{max,l} + \varphi_{max,r}}{2}\right|, \quad (16)$$

where $R_{1,2}(\varphi_{max})$ expresses the cross-correlation coefficient. $\varphi_{max,l}$ and $\varphi_{max,r}$ represent the left and right endpoints of the interval where the cross-correlation function takes its maximum value, respectively. The maximum value exists in an interval because binarized signals may show successive step-shaped results when solving for circular cross-correlation [see Fig. 7(b3)]. As the value of corner features related to the inter-layer criterion decreases, $\Delta_A$ decrease accordingly while $\Delta_\varphi$ increase accordingly. Based on the results in Fig. 7 (b), we suggest using $\Delta_A > 0.75$ and $\Delta_\varphi < 20°$ as the judgment criterion to filter out false corners from noise.

For the discrete sampler in the application, the sampling signals can be represented as $g_1(i)$ and $g_2(i)$, with the same length of 16 and each element represents the grayscale value of a sampling point (binarized). The circular cross-correlation function between two discrete signals can be calculated as

$$R_{1,2}(k) = \sum_{i=0}^{15} g_1^*(i) \cdot g_2^*(i+k), \quad k \in [0, 15], \quad (17)$$

where $g_1^*(i)$ and $g_2^*(i)$ represent the periodic extension of $g_1(i)$ and $g_2(i)$. At this point, the inter-layer corner response value can still be defined as $\Delta_A = R_{1,2}(k_{max})$ and $\Delta_\varphi = \left|\frac{\varphi_{max,l} + \varphi_{max,r}}{2}\right|$.

### 3.4. Rapid corner filtering

In Section 3.3, the intra-layer and inter-layer criterion for determining a CMP corner are obtained, and four corner feature response quantities are defined: $\Delta_{f_{12}}$, $\Delta_{f_{23}}$, $\Delta_A$, and $\Delta_\varphi$. In practical use, to avoid time-consuming mathematical calculations and meet the applicable requirement of real-time performance, the numerical analysis results of Section 3.3 can be used as a theoretical basis to design a simplified version of the fast corner filter.

*3.4.1 Simplified Intra-layer criterion*

1) Let the number of changes of black-and-white transitions in sampling sequence $k$ be $N_k$. A candidate corner needs to meet

$$N_k = \sum_{i=1}^{16} n_k(i) = 4, \quad (18)$$

where

$$n_k(i) = \begin{cases} 1, & g_k(i) \neq g_k(i+1) \\ 0, & g_k(i) = g_k(i+1) \end{cases}, k = 1, 2. \quad (19)$$

$g_k(x^*)$ denotes the x-th element (binarization intensity of pixels) in the sampling sequence $k$, which satisfies $x^* = x \mod 16$ (to describe a circular sequence). This criterion resembles Bok's approach of selecting candidate corners using sign-changing indices [37]. Besides, practices show that although the corrosion region less blurs the outer sampling circle, it is more likely to cover the excess blocks since it is far from the sampling center. Therefore, we added a cropping strategy to the calculation of the outer sampling sequence $g_1$, which can be expressed as

$$n_1(i) = \begin{cases} 1, & g_1(i) \neq g_1(i+1) \cap g_1(i) \neq g_1(i+2) \\ 0, & g_1(i) = g_1(i+1) \end{cases}. \quad (20)$$

Compared to Equ. (19), Equ. (20) adds tolerance for noise. On the one hand, we ensure that both sampling circles have the region feature of the four sectors in the CMP's corner model. On the other hand, we filter out jump points with a spacing equal to one pixel, which cuts out small pseudo sectors (see the outer sampling circle in Fig. 5).

2) For inner sampling sequence $g_2$, Let the length difference between black and white sector pairs be $\delta_b$ and $\delta_w$. Quantify the overall difference as $\delta_2$. A candidate corner needs to meet

$$\delta_2 = max(\delta_b, \delta_w) < \delta_{th}, \quad (21)$$

where $\delta_{th}$ is the selected threshold. From our experiment, 5 is selected as the threshold. The function of $max(\delta_b, \delta_w)$ is to replace the response quantities $\Delta_{f_{12}}$ and $\Delta_{f_{23}}$. From Fig. 6(c) and Fig. (d), the values of $\Delta_{f_{12}}$ and $\Delta_{f_{23}}$ are determined by the azimuth and length deviations based on the black sector. The length deviation represents the length difference between two vectors belonging to the black sectors, while the azimuth deviation represents the relative length between the centerline of the black sectors. In practical applications, both azimuth deviation and length deviation occur simultaneously. When using the white sector as the benchmark, the azimuth deviation can be explained as the difference in the length of the white sectors. Therefore, these two types of deviation can be uniformly described as the size difference $max(\delta_b, \delta_w)$ between sectors of the same color. The importance of this matter lies in avoiding time-consuming discrete Fourier transform. Practices show that the simplified intra-layer criterion has increased the computational speed by over 20 times.

*3.4.2 Simplified Inter-layer criterion*

In Section 3.3.2, we have proposed $\Delta_A$ and $\Delta_\varphi$ to describe the inter-layer corner feature. However, when only $\Delta_\varphi$ changes while $\Delta_A$ remains unchanged, although the maximum value of the cross-correlation function remains unchanged, the decrease of $R_{1,2}(0)$ is proportional to the increase of $\Delta_\varphi$ [see Fig. 7(b1)]. From Fig. 7(b2) and Fig. 7(b3), the change of $\Delta_A$ could also be reflected in the decrease of $R_{1,2}(0)$. Since the variations of $\Delta_A$ and $\Delta_\varphi$ are simultaneous in applications, we can use only the reduction of $R_{1,2}(0)$ to approximate the symmetry and similarity between the two layers of sampled signals.

Let the variance quantization of the two sampling sequences $g_1(i)$, $g_2(i)$ be $D_{1,2}$. A candidate corner needs to meet

$$D_{1,2} = \sum_{i=1}^{16} g_1(i) \oplus g_2(i) < D_{th}, \quad (22)$$

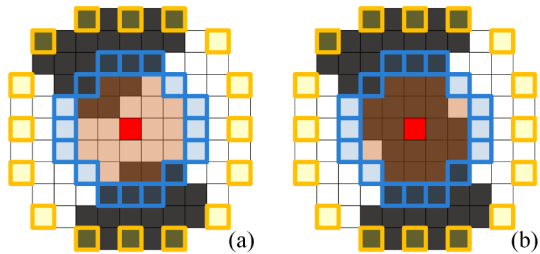

**Fig. 8.** Comparison of two features with similar responses in the proposed sampler. (a) Corner feature. (b) Stripe feature.

where $\oplus$ denotes XOR operation, $D_{\text{th}}$ is the selected threshold. 5 is chosen as the threshold according to our experiment. The advantage of using $D_{1,2}$ to approximate the inter-layer corner feature is that only a 16-bit XOR operation is required for each sampling center. The simplified intra-layer criterion can avoid the calculation of 16 times cross-correlation coefficients in Equ. (16) and increase the computational speed by more than 10 times.

### 3.5. Corner detection and refinement

Complete corner detection still faces an unresolved challenge: the above judgment criteria cannot distinguish a CMP corner from the center points of a stripe. Fig. 8 shows the detection effect when the sampling centers are located at a CMP corner and the center of a stripe, respectively. Both cases exhibit the same form judging only from the sampling sequence of the two sampling circles. It means relying solely on corner feature responses cannot distinguish between these two situations. This issue has also been discussed in the work of Bennett *et al.* [36]. Inspired by this research, we notice that the main difference between the two comes from the corrosion region. Although the discussion in Section 2.2 demonstrates that the corrosion region cannot provide sufficient credible details, we can use the overall characteristics of the entire area to reflect the level of corrosion: When the corrosion degree exceeds a certain value, it is considered that the sampling center may belong to a stripe feature rather than a corner feature.

Let the number of black pixels be $n_R$ and the number of white pixels be $21 - n_R$ in corrosion region. Let the numbers of black and white pixels in outer sampling sequence $g_1(i)$ be $n_1$ and $16 - n_1$, respectively. When the corrosion degree $Cd$ meets the requirement

$$Cd = max(n_1 - n_R, n_R - n_1 - 9) < Cd_{th}, \quad (23)$$

the current candidate corner can be regarded as a CMP corner. Equ. (23) requires that the difference in the number of pixels of the same color between the outer sampling sequence and the corrosion region should not exceed the threshold $Cd_{th}$. For example, the corrosion degree $Cd$ of the two features in Fig. 8 can be calculated as 1 and 7, respectively. The larger the value of $Cd$, the greater the degree of blurring or uniformity within the corrosion region, and the more likely the sampling point is to be inside a stripe. The outer sampling circle is used because a large circle can better cover the outside of stripes, thus making the stripe feature sensitive to the response of corrosion degree. From our experiment, 4 is the preferred threshold value.

The proposed corner filters using the conditions of Eqs. (18), (21), (22), and (23) can effectively filter out the candidate domains from the tactile images that have the potential to belong to the corners. Due to the characteristics of CMP, candidate domains are represented as clusters of connected domains located near the corners, as shown in Fig. 3(c). Bennett

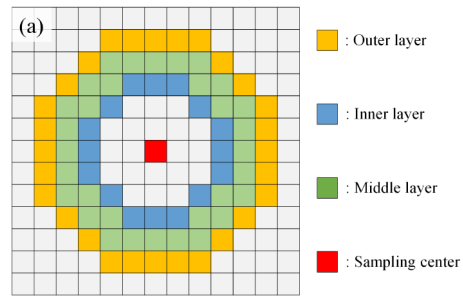

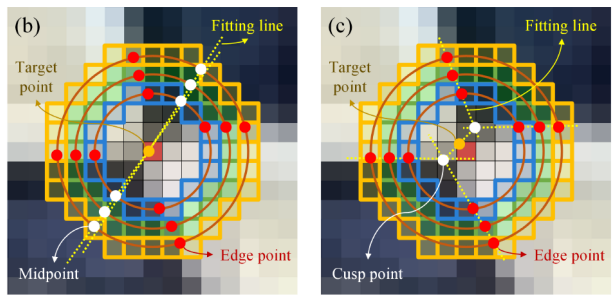

**Fig. 9.** Sampler for corner refinement. (a) Sampler design. (b) Sub-pixel localization based on midpoints. (c) Sub-pixel localization based on edge points.

*et al.* argued that a corner detector should better provide process information rather than a Boolean-type response value [36]. Although the local threshold binarization method is used in this article, we approve that assigning an explicit response to each pixel in the acquired candidate domains can help us to locate the specific position of the corners accurately. We define the symmetrical response $R$ as

$$R = \delta_{\text{th}} - \delta_2, \quad (24)$$

where the definition of $\delta_2$ and $\delta_{\text{th}}$ comes from Equ. (21). The inner sampling sequence is considered because it is closer to the sampling center. The symmetrical response $R$ reflects the symmetry of the corner sectors. The closer the sampling center to the corner, the smaller the value of $\delta_2$ and the higher the response value $R$. After assigning a response value to each pixel in the candidate domains, we use the standard non-maximum suppression to discard low responses in a small neighborhood around each pixel in the response image, thereby retaining only the candidate corners with the highest response value.

At this point, the integer pixel coordinates of each candidate corner point are determined. Although the discussion in Section 2.1 indicates that the accuracy requirement for CMP-based visuotactile sensing is not high, we still provide the refinement technique suitable for the feature of CMP corner to avoid incorrect positioning. Common refinement techniques include methods based on saddle-point [32], symmetric [25], and iterative (e.g., the function *cornerSubPix* in OpenCV). However, due to the presence of corrosion regions, such methods are not the optimal choice in situations where the geometric distortion caused by contact is large.

- For saddle-point-based methods, the high noise in the corrosion region could cause the saddle-point approximation result hard to predict.
- For iterative-based methods, when the corrosion degree is significant, the refinement result may approach one of the sharp cusps of the sectors (located on the periphery of the corrosion region).
- For symmetric-based methods, since the CMP corner no longer possesses symmetry, the refinement results may be more biased toward

the interior of the black or white sectors for corners with large azimuth offsets.

Considering the above factors, we propose a refinement approach based on the midpoints. Corner refinement uses the three-layer sampler, as shown in Fig. 9 (a). Since the corner refinement process handles fewer data than corner detection, we can use more sampling information. Compared to the sampler for corner detection, the three-layer sampler adjusts the shape of the outer sampling circle. It adds a middle layer for interpolating information between the inner and the outer circles. The purpose of such modification is to ensure continuity in both the radial and lateral directions of the sampling.

Fig. 9 (b) shows the process of refining corners using this method. Note that we are now considering the original tactile image rather than the binarized one. By searching for the edge points at the boundary of sectors in each layer of the sampler, the midpoints of two edge points in the same sector can be used as the sampling point for the sector centreline. Take the black sector as an example. We can fit the midpoint of three layers to obtain two sector centrelines and use their intersection points as the refinement result. The specific process includes:

- In the binary image, find the integer pixel coordinates of all edge points in the three-layer sampler.
- Return to the original tactile image, use Li's sub-pixel correction method [16] or Gioi's subpixel edge detection method [41] to calculate the subpixel coordinates of edge points based on the obtained integer coordinates.
- Use the same sector's edge points to calculate the sector's midpoints, and use the linear or quadratic function equation to fit three midpoints to calculate the expression of a centreline.
- Calculate the intersection point of the centreline of two black sectors. The intersection point is the refined corner position.

In addition, in some cases, the calculated two centrelines may intersect within the black and white sectors or even be parallel. If the refined position is far from the original place (for example, take 2 pixels as the threshold), we suggest using another edge-point-based method to refine the corners [see Fig 9(c)]:

- Use the linear or quadratic function equation to fit three edge points and calculate the expression of the edge line.
- For the same sector, calculate the intersection point of two edge lines as the cusp point.
- Calculate the midpoint of the line connecting two cusp points. The midpoint is the refined corner position.

The core of the corner refinement method based on the midpoints or the edge points mentioned above is the boundary feature of the CMP corner. By utilizing the edge points obtained from the multi-layer sampler, corner position can be obtained through boundary fitting. Since this approach does not rely on pixel intensity in the corner neighborhood, it can effectively avoid the influence of corrosion regions.

## 4. Experiments

In this section, we used the Tac3D 3.0 sensor [17] to validate the robustness and efficiency of our proposed detection method through comparative experiments. The reconstruction and visualization of dense 3D contact deformation achieved by this approach were also demonstrated. The structure of Tac3D is shown in Fig. 10(a). The overall size was 90 mm × 70 mm × 90 mm. Its body was made by 3D printing, and an RYS-1200-Farbe-global camera (1600×1200 pixels) was selected to capture the contact deformation of the marker pattern. The soft elastomer, camera, LED light source, and metal mirrors were all fixed on the housing. The Tac3D sensor adopted the virtual binocular vision system (VBVS) [42] by means of the optical path structure shown in Fig. 10(b). Through clever optical circuit design, it could achieve binocular stereo vision using just one camera. Therefore, the Tac3D sensor enabled binocular measurements with high synchronization and had good compactness simultaneously.

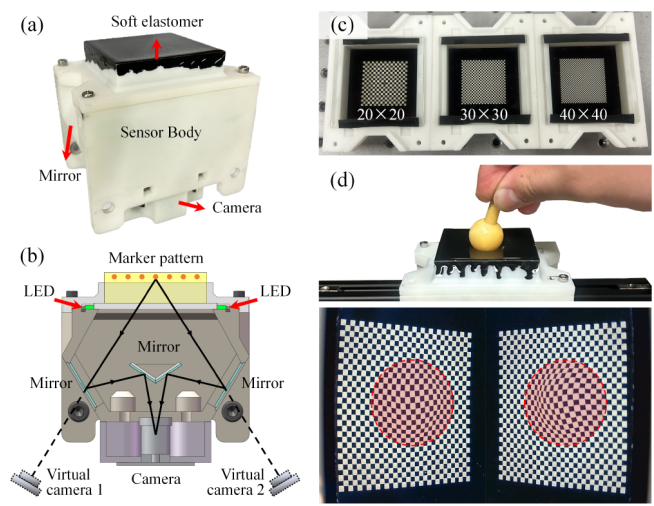

**Fig. 10.** Experimental setup for robustness and efficiency evaluation. (a) The structure of Tac3D 3.0 sensor [17]. (b) The virtual binocular vision system used in Tac3D [17]. (c) The detachable soft elastomers with different densities of CMP. (d) Example of an experiment. The contact between an object and the soft elastomer could be reflected in tactile images.

The soft elastomers of Tac3D were made of Shinbon silica gel with a base-to-curing agent ratio of 1:1 to obtain an appropriate hardness. They were designed to be removable to be easily replaced during experiments. We made three elastomers with different densities of CMP attached to them. All CMP sizes were 32mm × 32mm, and the sizes of their marker array were 20×20, 30×30, and 40×40, respectively [see Fig. 10(c)]. We selected different test objects (ball, torus, cube, rib, and wedge) to contact the soft elastomer of Tac3D. When the marker pattern deformed during contact, real-time photography was taken using the camera built into the sensor [see Fig. 10(d)]. By using the method proposed in this article and other baselines for detection on the same tactile image, the performance of different approaches could be fairly compared. All algorithms were implemented by C++ OpenCV, and ran on a computer equipped with an Intel i7-12700H processor at 2.30 GHz.

### 4.1. Real-World robustness evaluation

This section evaluates the robustness of the proposed method in CMP-based visuotactile sensing tasks. The dataset used includes a total of 600 tactile images and 580000 valid corner markers. We used five different shapes of objects and three different densities of CMP (size of marker array: 20×20, 30×30, and 40×40). When the test object comes into contact with the Tac3D sensor equipped with the corresponding CMP, pressing, shearing, or twisting are performed in different contact modes. For the same object and CMP with the same density, 20 binocular tactile images are collected. Each image contains 800, 1800, or 3200 valid corners.

Three reference methods were used to run for the same task to make a comparison with the proposed method: the ChESS detector [36], Zhang's method [38], and Shi-Tomasi's method [43] (using the function *goodFeaturesToTrack* in OpenCV). ChESS is a response graph-based circular detector that can achieve outstanding detection results when the threshold is selected appropriately. The method proposed by Zhang *et al.* introduces spectrum analysis in the ChESS detector, which is closest to the detection method proposed in this article. Shi-Tomasi's method is a universal corner detection method, and we chose it to demonstrate the necessity of customizing the detection method for visuotactile perception.

**Table 2**
Parameter selection for different methods.

| Method | Variable parameter (Image grayscale: 0~255) | Values selection for different CMPs | | |
|---|---|---|---|---|
| | | $20^2$ | $30^2$ | $40^2$ |
| Proposed | Window size of adaptive binary | 40×40 px$^2$ | 30×30 px$^2$ | 20×20 px$^2$ |
| | $\delta_{th}, D_{th}, Cd_{th}$ | 5, 5, 4 | | |
| ChESS | Threshold of overall response | 0.35 | 0.50 | 0.70 |
| | Radius of sample-ring | 5 px (16 sample points) | | |
| Zhang's | Threshold of $|f_1|-|f_2|$ | 1.80 | 2.15 | 2.45 |
| Shi-Tomasi's | maxCorners | 800+168 | 1800+248 | 3200+328 |
| | minDistance, qualityLevel | 8, 0.01 | | |

**Table 3**
Success rate (%) of the proposed method and reference methods (ChESS [36], Zhang et al [38], and Shi-Tomasi's [43]).

| Algorithm | Test object & Size of marker array | | | | | | | | | | | | | | | Average |
|---|---|---|---|---|---|---|---|---|---|---|---|---|---|---|---|---|
| | Ball | | | Torus | | | Cube | | | Rib | | | Wedge | | | |
| | $20^2$ | $30^2$ | $40^2$ | $20^2$ | $30^2$ | $40^2$ | $20^2$ | $30^2$ | $40^2$ | $20^2$ | $30^2$ | $40^2$ | $20^2$ | $30^2$ | $40^2$ | |
| **Proposed** | **100.0** | **100.0** | **100.0** | **100.0** | **100.0** | **97.5** | **100.0** | **100.0** | **92.5** | **97.5** | **97.5** | **92.5** | **97.5** | **95.5** | **92.5** | **97.53** |
| ChESS | 77.5 | 12.5 | 20.0 | 90.0 | 10.0 | 10.0 | 92.5 | 27.5 | 25.0 | 75.0 | 22.5 | 22.5 | 77.5 | 22.5 | 35.0 | 41.33 |
| Zhang's | 92.5 | 2.5 | 0.0 | 87.5 | 12.5 | 0.0 | 87.5 | 20.0 | 0.0 | 90.0 | 22.5 | 0.0 | 90.0 | 22.5 | 0.0 | 35.17 |
| Shi-Tomasi's | 0.0 | 0.0 | 0.0 | 0.0 | 0.0 | 0.0 | 0.0 | 0.0 | 0.0 | 0.0 | 0.0 | 0.0 | 0.0 | 0.0 | 0.0 | 0.0 |

This evaluation used three indicators to reflect the effectiveness of different methods: *Average False Positive (AFP)*, *Average False Negative (AFN)*, and *Success Rate (SR)*. Among them, *AFP* and *AFN* represented the average number of false and missed corners in the same tactile image, respectively. The larger the value of *AFP* and *AFN*, the worse the detection effect of the relevant method. When *AFP* or *AFN* exceeded 1 (i.e., not all corners are correctly detected), it was considered that the detection of that frame of tactile image failed. *SR* was used to describe the algorithm's success rate in handling the test samples in the used dataset. The *AFP* and *AFN* indicators are only used to visually demonstrate the differences in different methods, while the *SR* indicator can better reflect the robustness of the processes in practical applications. To ensure fairness in comparison, we manually selected the optimal parameter combination for each of the four algorithms under different contact conditions, as shown in Table 2. The standards for parameter selection were such that the *AFP* and *AFN* indicators were small, along with a small gap between *AFP* and *AFN*. Therefore, the detected corners are neither too many nor too few.

The *Average False Positive* and *Average False Negative* for each method are summarized in Fig. 11. The proposed method achieved the best detection results in all cases. On the experimental dataset, our algorithm held the *AFP* and *AFN* averaged 0.020 and 0.043, with an average of no more than 0.05 errors in a tactile image containing thousands of corners. In addition, the AFP and AFN of our method did not vary as significantly as the other three methods. When the array size of CMP is 20×20, both the ChESS detector and Zhang's method could show more satisfactory performance, but the errors appeared significantly increase with the growth of CMP's size. The reason is that the single-layer circular detectors could not identify the distorted corners with weak feature responses. Shi-Tomasi's method obtained similar detection results in the cases of 20×20 and 30×30 array sizes, but a surge in errors occurred at the 40×40 array size. It implied that the corrosion degree had reached the upper limit of recognition of the universal detector as the marker density increased. The above discussion demonstrates that our method is less affected by the variation of CMP feature density than the other three methods and has good stability in the 3D deformation measurement with even high marker densities.

The experimental results shown in Fig. 11 also indicate that different contact states could have impacts on the detection. The ChESS detector and Zhang's method performed worse in smooth contact (ball and torus surfaces)
12

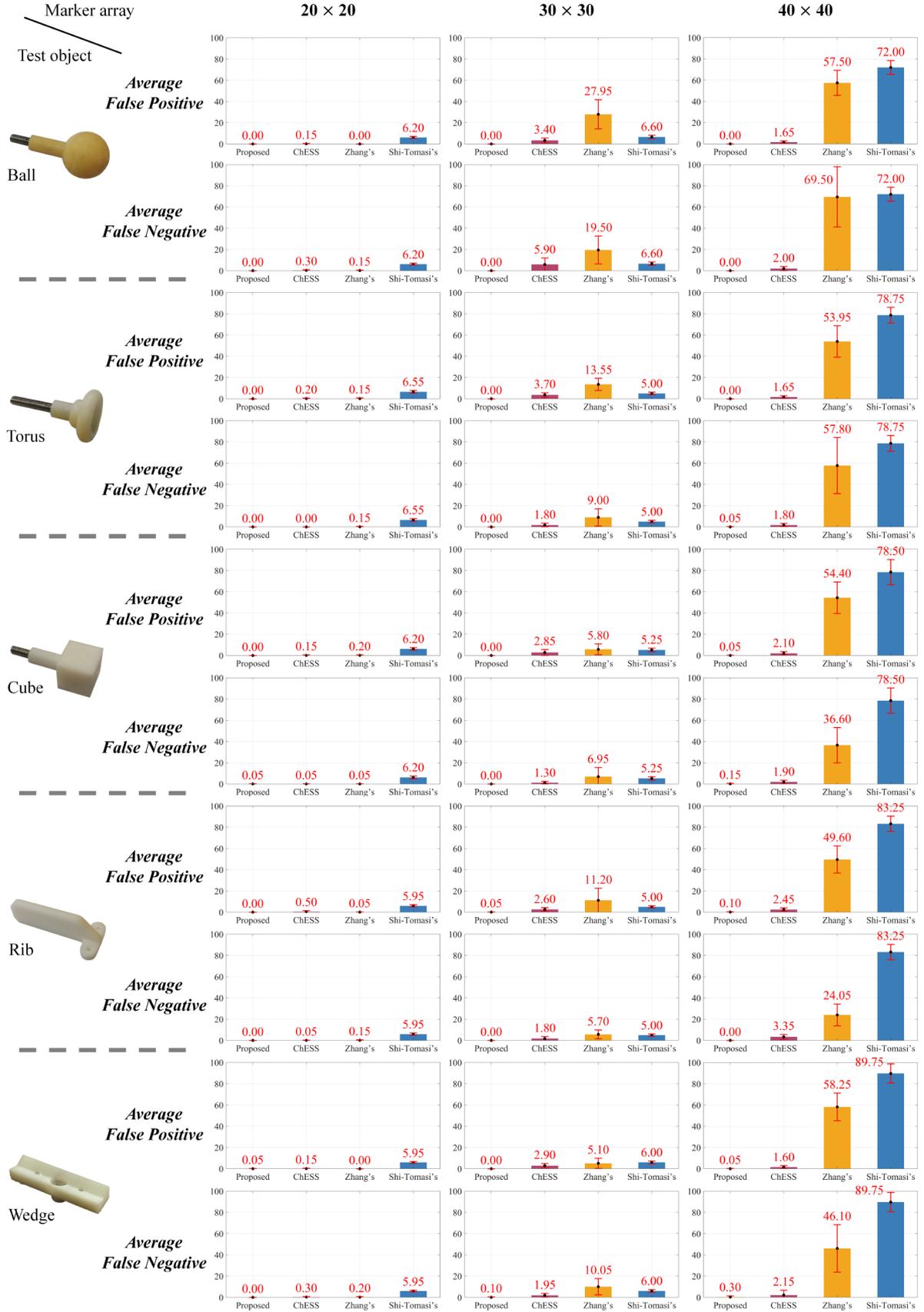

**Fig. 11.** Evaluations of real-world robustness of the proposed method and reference methods (ChESS [36], Zhang *et al* [38], and Shi-Tomasi's [43]) in practical applications, using five different test objects and three CMPs with different densities. *Average False Positive* and *Average False Negative* metrics were used to emerge performance differences.

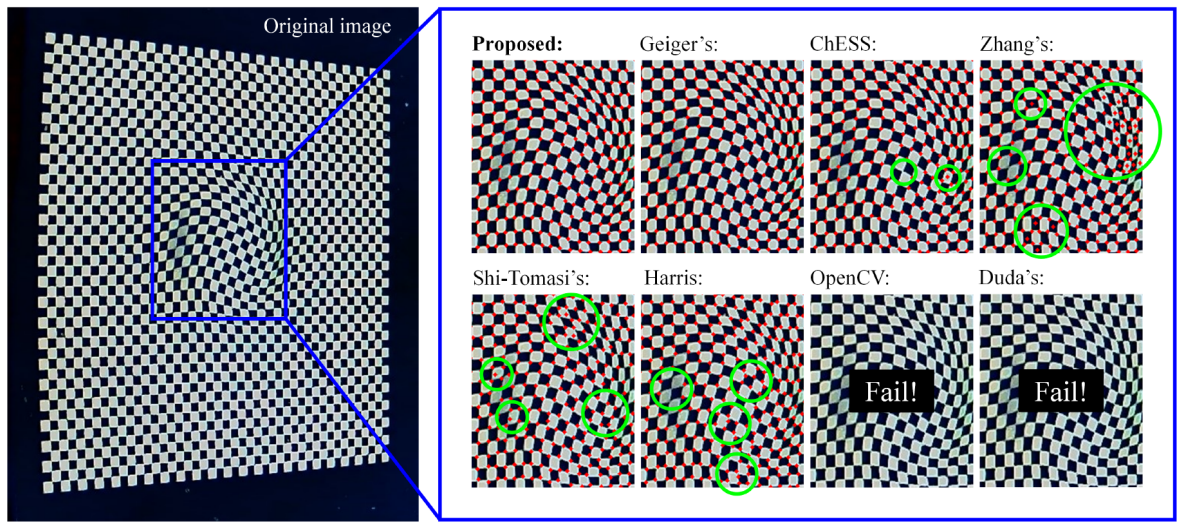

**Fig. 12.** Feature detection results provided by the proposed method and reference methods (Geiger's [20], ChESS [36], Zhang et al [38], Shi-Tomasi's [43], Harris [44], the function *findChessboardCorners* in OpenCV, and Duda et al [21]).

compared to the sharp contact (cube, rib, and wedge surfaces). In contrast, Shi-Tomasi's method exhibited the opposite characteristics. For the former case, the marker units of CMP were sheared by curved surfaces and generated tangential squeeze, reflected in the symmetry destruction of the corners. It made the methods (ChESS and Zhang's) that were based on symmetry considerations prone to false positives. In the latter case, the marker units of CMP were stretched tangentially due to the compression of sharp surfaces, resulting in connectivity failure at the corners. In this case, the universal detection approach represented by Shi-Tomasi's method is prone to detecting two positive results. Therefore, due to the simultaneous consideration of these two factors in the specific case of CMP-based visual-tactile sensing, the robustness of the proposed method is significantly improved compared to the existing techniques.

Compared to the latter two reference methods, the performance of the ChESS detector was close to that of the proposed method. On average, the ChESS detector could reach approximately 1.69 errors in a single tactile image. However, although the average number of mistakes in the ChESS detector is small, this does not mean that the reliability of the ChESS detector could reach a similar level as the proposed method. It is because any false or missed corners indicate detection failure. Table 3 shows the *SR* indicator for four different approaches. From the results, the success rate of the ChESS detector is closer to that of Zhang's method, which is only 0.42 times that of our method. Therefore, our algorithm's efforts in handling false positive and false negative responses are more evident in robust performance indicators.

On the basis of previously considered algorithms, we have added some new reference methods: the Harris detector [44], Geiger's method [20], the function *findChessboardCorners* in OpenCV and Duda's method (using the function *findChessboardCornersSB* in OpenCV) [21]. In the qualitative experimental example shown in Fig. 12, both the proposed method and Geiger's method correctly detected all corners. In contrast, the ChESS detector occured both anomalies and omissions in the middle positions, while the algorithm proposed by Zhang *et al.* gave a large number of outliers in the area with obvious deformation and extrusion. The universal detectors (Harris's and Shi Tomasi's) also displayed poor performances.

Corner omissions occurred at locations with uneven lighting, while a large number of abnormal results were found around the corners that had a significant corrosion degree (the separated sharp cusps are detected as two corners). For these complete visual calibration methods, due to the significant geometric distortion of the pattern units, the OpenCV method based on contour extraction, and Duda's method based on Laplace transform could not run successfully. Overall, existing detection methods could result in many errors and might fail to detect the tactile images with dynamic deformation, while our method outperformed these competing algorithms in visuotactile sensing.

It is worth noting that Geiger's method also showed good detection performance, which seems to indicate that it has substitutability with the detection method proposed in this article. However, this performance is to some extent dependent on the complex judgment benchmark, which makes this method, time-consuming and unable to meet real-time requirements. The details are explained in Section 4.2.

*4.2. Computational efficiency evaluation*

To further evaluate the computational efficiency, we compared the average wall-clock execution time calculated by the different algorithms after processing the same VGA-resolution tactile image. Each method ran 1000 times under the same hardware condition, and a single tactile image contained 1936 corner features to be detected. In this section, we considered four algorithms: the ChESS detecter, Shi Tomasi's method, Zhang's method, and Geiger's method. All procedures were implemented through C++. Without special instructions, the code of different methods was ensured to work at the same optimization level as far as possible.

The time spent results are shown in Table 4. Compared to Geiger's method with similar detection results (according to), the proposed method can save nearly 800% of the time. Although Geiger's approach could achieve ideal detection results, it failed to run in real-time as our method. Compared with Zhang's method, which also considers the analysis of amplitude-frequency characteristics, our algorithm could avoid complex Fourier transform operations using the simplified fast detector. Compared

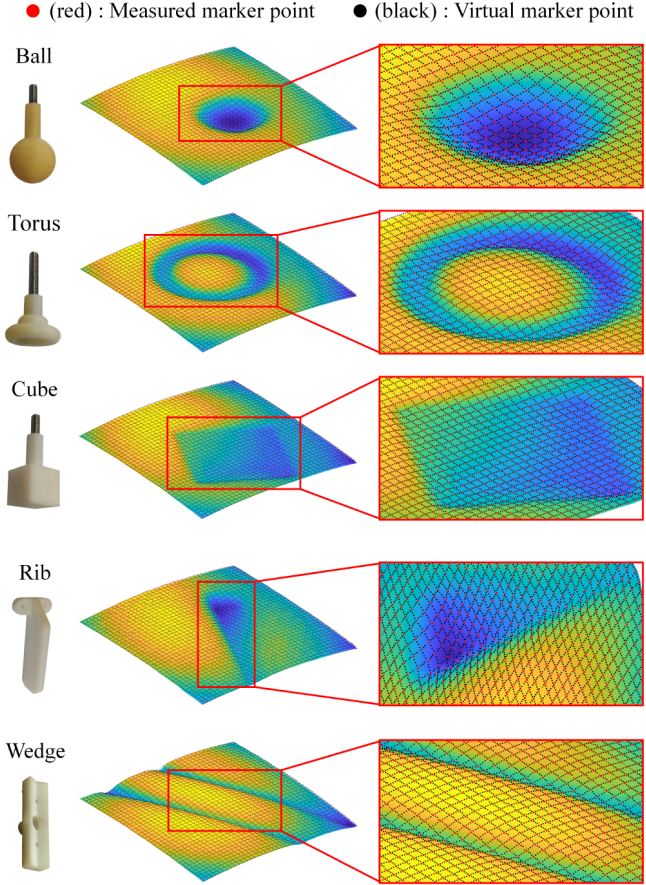

● (red) : Measured marker point   ● (black) : Virtual marker point

Ball

Torus

Cube

Rib

Wedge

**Fig. 13.** Dense 3D deformation measurement based on the proposed method. Five different test objects were used. The red dots represent the measured marker points, and the black dots represent the virtual marker points.

**Table 4**
Time spent to perform the proposed method and reference methods (ChESS [36], Zhang *et al* [38], Shi-Tomasi's [43], and Geiger's [20]).

| Algorithm | Average wall-clock time (ms) |
|---|---|
| Proposed | 8.351 |
| Proposed with OpenMP | **4.951** |
| ChESS | **5.041** |
| Shi-Tomasi's | 7.029 |
| Geiger's | 653.6 |
| Zhang's | 1313.5 |

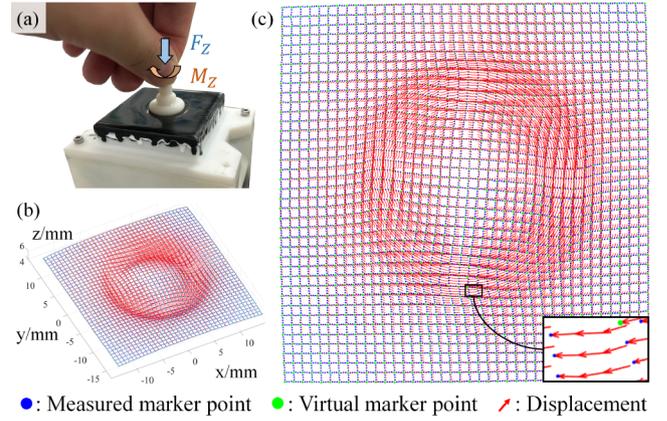

●: Measured marker point   ●: Virtual marker point   ↗: Displacement to the ChESS detector and the universal corner detector Shi Tomasi's, which have significant advantages in detection efficiency, our detection method used more sampling information and complex judgment indicators, resulting in a relatively slow running speed. However, this relatively small difference in speed was achieved with a higher success rate of 97.53% (Bennett's and Shi-Tomasi's had success rates of 41.33% and 0%, respectively). It means that the proposed method could achieve better detection performance while still meeting the real-time requirement in terms of operational efficiency.

**Fig. 14.** Demonstration of dynamic contact sensing. (a) Squeeze and twist on the contact surface using a torus. (b) Reconstruction of dense 3-d displacement field (c) Orthographic projection view of the displacement field.

Note that the criteria for corner feature response in the proposed detection method are not strictly sequential, which means that parallel operations could effectively improve the computational speed. By parallel processing multiple pixels using OpenMP, the proposed method could achieve a running speed similar to the ChESS detector. Therefore, this detection method could handle tactile images containing 1936 feature points at approximately 200Hz. Such performance is far enough to meet the requirements of deformation measurement.

*4.3. Dense 3-d deformation measurement*

By combining with the work of Li *et al.* [16], dense 3-d contact deformation measurement was achieved. In the new detection process, the feature detection algorithm proposed in this article replaces the marker recognition step based on polygon fitting in [16], but retains the construction step of topological connection relationship based on contour information. By performing corrosion operations on the neighborhood of each corner, the marker units could be separated more accurately while reducing the loss of original details in high-density CMP. We also used the virtual marker points mentioned in [16] to perform discrete sampling on the edge lines of CMP. Three virtual marker points were selected between every two measured marker points. They could be used to characterize the deformation together with the measured marker points obtained using the proposed method.

Fig. 13 shows the contact deformation visualization of five typical objects. Our method could reconstruct the contact deformation with clear details, whether under smooth or sharp contact characteristics. Due to the introduction of the detection algorithm proposed in this article, we could practically apply a 40 × 40 CMP in the Tac3D sensor to achieve a high-density representation of an average of about 10.7 effective markers per square millimeter. It enabled us to attain ultra-fine contact deformation sensing that is difficult to achieve by other visuotactile sensors based on the marker displacement method.

Besides, the dense 3-d deformation measurement we achieve should be understood as something other than 3-d morphology perception. As shown in Fig. 14, we used a torus-like object to contact the Tac3D sensor and applied positive pressure and torque. The method implemented in this article can clearly record the process of contact deformation in the form of a dense displacement field. Unlike static contact geometry, our approach could realize retro-graphic sensing of the dynamic contact process. It is a unique function that cannot be achieved by the photometric stereo method, the other dominant method in visuotactile sensors.

*4.4. Discussion and extension*

The experiments in Section 4.1 and Section 4.2 demonstrate that the proposed method has advantages in real-time and robustness, making it suitable for CMP-based visuotactile sensing with high information density. Compared with the reference methods, we use a double-layer circular sampler and propose corresponding intra-layer and inter-layer response indicators based on numerical analysis. Due to fully considering the properties of the corner model under dynamic deformation, the proposed method exhibited the best performance in resisting geometric distortions caused by contact behaviors such as squeezing and pulling. Finally, thanks to the effective application of this method in marker recognition, we successfully achieved dense 3-d contact deformation measurement.

Compared with the mainstream digital image correlation (DIC) and optical flow method, the proposed approach for deformation measurement does not rely on tracking the deformation process. Using contour information between markers in CMP, we can construct a fixed topological connection relationship and assign the marker points an order number that does not change with the state. The above characteristics enable us to calculate the deformation relationship between any non-adjacent frames of images without the need for real-time tracking of the marker movement. In Fig. 14, we only used the 3rd, 5th, 9th, and 12th frame images to calculate the displacement field after the contact occurred. Suppose the digital image correlation method or optical flow method is used. In that case, the measurement may be invalid since the time interval is too large to judge the similarity of the sub-regions. The above advantages allow our method to be applied to scenes with more complex working conditions, such as calibrating structured light cameras. In addition, our process benefits from analysing the amplitude-frequency characteristics and cross-correlation of sampled information, and such an approach is also expected to expand to other types of marker patterns (such as the triangular chessboard [45]).

However, in order to apply to tactile images with significant geometric distortion, our method focuses on loose and conservative judgment indicators. Meanwhile, due to the corrosion regions in CMP, our demand and attention for accuracy are lower than those works in the field of visual positioning. Therefore, the proposed in this article may not be suitable for camera calibration tasks with good imaging quality and large size of marker patterns.

## 5. Conclusion

This article proposes a feature detection method for dense 3-d contact deformation measurement. Our study starts by analyzing the characteristics and requirements of CMP-based visuotactile sensing tasks. Based on the proposed feature model, we design a multi-layer circular sampler to provide feature response to the CMP corner. It relies on the intra-layer criterion based on amplitude-frequency characteristics and the inter-layer criterion based on circular cross-correlation, which can effectively suppress different types of errors. Through comparative experiments on the Tac3D sensor, we evaluated the performance of the proposed method and demonstrated its optimal performance in actual practical use. Compared with existing feature detection methods, the proposed method exhibits more prominent real-time and reliability advantages in visuotactile sensors.

The subsequent research will focus on further refining the method of visuotactile sensing based on continuous marker patterns. On the basis of the proposed method, the understanding and utilization of contour information will be further refined to achieve more comprehensive and applicable contact deformation measurements in more complex scenes. In addition, we are about to explore the applications of dense 3-d deformation visualization and extend our research contributions to human-computer interaction, interface science, and other fields.

## Acknowledgements

This work was supported in part by the National Natural Science Foundation of China under Grant 52175017, in part by the Joint Fund of Advanced Aerospace Manufacturing Technology Research under Grant U2017202, in part by a grant from the Institute for Guo Qiang, Tsinghua University, and in part by the National Training Program of Innovation and Entrepreneurship for Undergraduates under Grant 202210003017.